\documentclass[letterpaper, 10 pt, conference]{ieeeconf}  

\IEEEoverridecommandlockouts                              

\usepackage{epsfig} 
\usepackage{amsmath} 
\usepackage{amssymb}  
\usepackage{multirow}
\usepackage{url}
\title{\LARGE \bf Beyond Isolated Heads: Multi-Overlapped-Head Self-Attention for Vision Transformers}

\author{Tianxiao Zhang$^{1}$, Bo Luo$^{2}$ and Guanghui Wang$^{3}$
\thanks{$^{1}$Tianxiao Zhang is with the Department of Computer Science, College of Staten Island, City University of New York, Staten Island, NY 10314, USA, and with the Department of Computer Science, The Graduate Center, City University of New York, New York, NY 10016, USA, and with the Department of Electrical Engineering and Computer Science, University of Kansas, Lawrence, KS 66045, USA.
        {\tt\small Tianxiao.Zhang@csi.cuny.edu}}%
\thanks{$^{2}$Bo Luo is with the Department of Electrical Engineering and Computer Science, University of Kansas, Lawrence, KS 66045, USA.
        {\tt\small bluo@ku.edu}}%
\thanks{$^{3}$Guanghui Wang is with the Department of Computer Science, Toronto Metropolitan University, Toronto, ON M5B 2K3, Canada.
        {\tt\small wangcs@torontomu.ca}}%
}

\begin{document}

\maketitle
\thispagestyle{empty}
\pagestyle{empty}

\begin{abstract}

Multi-Head Self-Attention (MHSA) is the cornerstone of Vision Transformers, allowing models to capture diverse feature representations by projecting tokens into independent subspaces. However, the standard MHSA strictly isolates these heads, preventing any information exchange during the attention computation itself. In this paper, we propose Multi-Overlapped-Head Self-Attention (MOHSA), a novel mechanism that replaces the hard division of attention heads with a soft, overlapping division. By allowing queries, keys, and values to partially overlap with those of adjacent heads, MOHSA fosters rich inter-head communication directly within the attention mechanism. We extensively investigate various overlap dimension scheduling strategies to identify optimal configurations. Comprehensive experiments in several Transformer architectures demonstrate that MOHSA outperforms standard MHSA on the CIFAR-10, CIFAR-100, Tiny-ImageNet, and ImageNet-1k datasets, offering a significant performance boost with negligible computational overhead. Code link: \url{https://github.com/ZTX-100/MOHSA}.

\end{abstract}


\section{Introduction}
Since the introduction of Transformer \cite{vaswani2017attention} from language to vision, Transformers have gradually dominated many vision tasks such as image recognition \cite{dosovitskiy2020image}\cite{liu2021swin}\cite{chen2023accumulated} and object detection \cite{carion2020end}\cite{ma2021miti}\cite{zhao2024ms}. Due to the global attention mechanism in Transformers, Vision Transformers could yield better performance in vision tasks when the training data is abundant and the training epochs are sufficient. Although the convergence rate of Vision Transformers is frequently slower than their convolutional counterparts, global information communication is one of the key elements for the success of Transformer models.

Multi-Head Self-Attention (MHSA) \cite{vaswani2017attention} represents that the queries, keys and values are split into different heads and the self-attention is computed in each head independently. Transformer models with Multi-Head Attention frequently yield better performance than those with Single-Head Attention. Thus, most Transformer models utilize Self-Attention with multiple heads for better representation learning and performance. Even though MHSA boosts the performance of the Transformer models, the relationships and interactions between different heads are rarely investigated.

The success of Transformer models lies mainly in the effective information exchange between different tokens so that each token can have a global view of the context information. Due to the superior performance of MHSA, most Transformer models utilize MHSA by default. However, the queries, keys and values are divided for each head without overlapping, and there is no information exchange when the attention is computed in each head. In other words, when calculating the attention in the current head, it does not have the information in other heads. Although the tokens will be processed by linear projections after attention, the information exchange is limited only to each token.

\begin{figure}[t]
\begin{center}
   \includegraphics[width=0.9\linewidth]{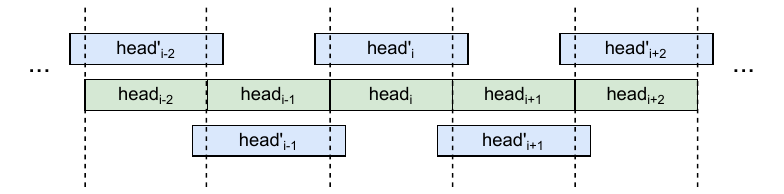}
\end{center}
   \caption{The proposed multi-overlapped-head method (blue) vs the original multi-head method (green). Instead of a hard division of the heads, our approach softly splits the heads by overlapping each head with its neighboring heads.}
\label{fig:heads_overlap}
\end{figure}

In this paper, we claim that information exchange during attention calculation in each head can improve the performance of Vision Transformers. This can be realized by overlapping queries, keys and values in each head with queries, keys and values of neighboring heads. For this purpose,
we propose Multi-Overlapped-Head Self-Attention (MOHSA) to improve the Multi-Head Self-Attention mechanism by overlapping the heads so that $Q$, $K$, and $V$ in each head overlap with $Q$, $K$, and $V$ of their neighboring heads when the attention is calculated, as illustrated in Fig.~\ref{fig:heads_overlap}. By overlapping the heads, the information in other heads could also be involved in the calculation of the attention in the current head. Since overlapping would slightly increase the dimension of the tokens after concatenating tokens from different heads, linear projections would decrease the dimension of the tokens to the original size. Moreover, we design various paradigms for overlapping ratios to investigate the best performance of our proposed approach. 

Our major contributions are summarized below.
\begin{itemize}\itemsep0em 

    \item We propose Multi-Overlapped-Head Self-Attention (MOHSA) and demonstrate that Vision Transformer models could be improved by overlapping the queries, keys and values of the current head with the queries, keys and values of adjacent heads when the attention is computed. Several Vision Transformer models are exploited on various datasets to demonstrate the effectiveness of our proposed method. 
    \item Various variants are proposed on the basis of overlap dimensions to thoroughly investigate the optimal performance of MOHSA. The proposed MOHSA could be integrated into Vision Transformer models to enhance their performance with negligible overhead.
    \item To the best of our knowledge, our work is the first to investigate the overlapping approach between different heads for MHSA when the attention is calculated.
\end{itemize}

\section{Related Work}

\subsection{Vision Transformers}

Vision Transformer \cite{dosovitskiy2020image} utilizes image patches which are embedded as tokens and a class token as token inputs for the Transformer encoder to recognize images. Swin-Transformer \cite{liu2021swin} reduces the computational cost by limiting the implementation of the attention calculation in each window and expands the view of the tokens by changing the windows. 
CaiT \cite{touvron2021going} and DeepViT \cite{zhou2021deepvit} investigate Vision Transformers with deeper layers. A multi-layer dense attention decoder was proposed in \cite{patel2024multi} for segmentation. MobileViT series \cite{mehta2021mobilevit}\cite{mehta2022separable}\cite{wadekar2022mobilevitv3} explore Vision Transformers in mobile-level applications. Some works \cite{ibtehaz2024acc}\cite{zhang2025depth}\cite{zheng2025iformer} introduce convolutions into the Vision Transformers to take advantage of both convolutions and Transformers. ViT-5 \cite{wang2026vit} advances the architecture of Vision Transformers for further performance enhancement.

\subsection{Attention Heads Interaction}

DeepViT \cite{zhou2021deepvit} explores a deeper-layer Vision Transformer by proposing re-attention that mixes the attention maps among all heads with a learnable matrix before multiplying with the values. The re-attention is similar to talking-heads attention \cite{shazeer2020talking}, which is originally employed for language tasks and also utilized by CaiT \cite{touvron2021going}. Talking-heads attention \cite{shazeer2020talking} applies learnable linear projections to heads in Multi-Head Attention before and after the softmax function to exchange information for the attention maps between heads in the attention module.

In contrast, we propose a method that enables communication between heads during attention computation by overlapping the heads with $Q$, $K$, and $V$. Although this approach introduces a slight increase in computation and parameters, it significantly enhances the performance of Vision Transformer models across various datasets.

\section{Approach}

\begin{figure*}[t]
\begin{center}
   \includegraphics[width=1.0\linewidth]{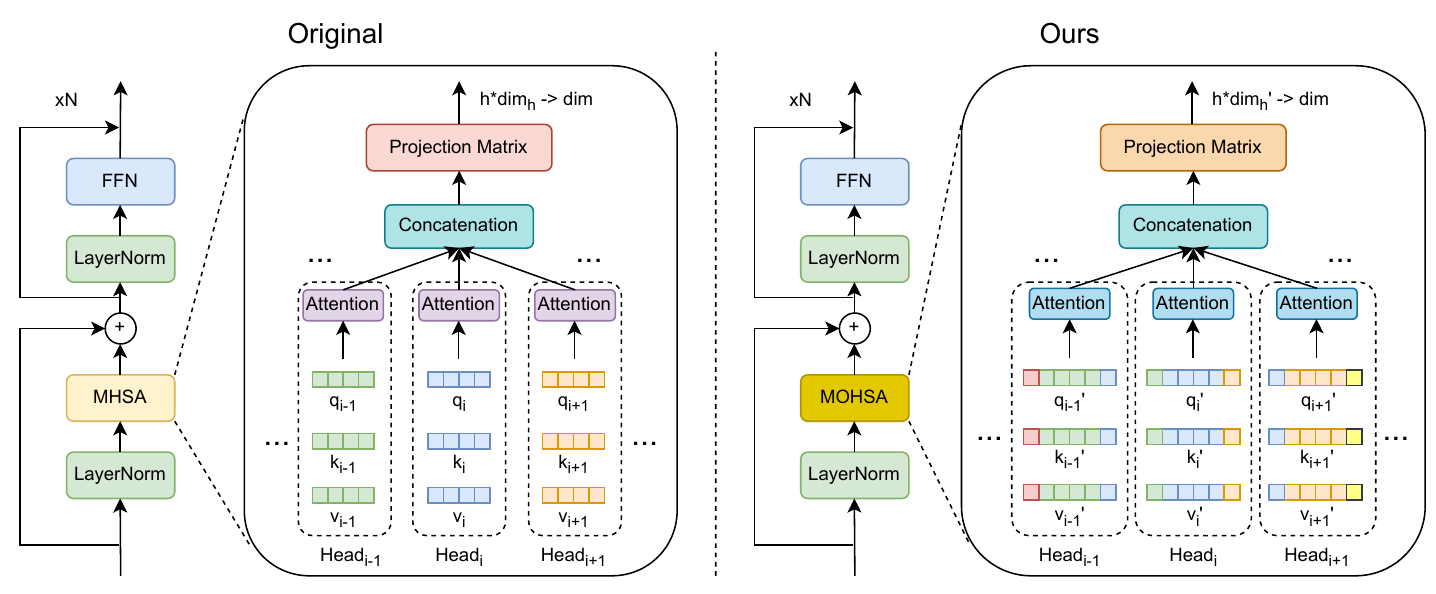}
\end{center}
   \caption{Our proposed Multi-Overlapped-Head Self-Attention. MHSA represents the original implementation of Multi-Head Self-Attention with hard division of heads and MOHSA indicates our proposed Multi-Overlapped-Head Self-Attention with soft division of heads. In the original Vision Transformer (left), $Q$, $K$, and $V$ are split for different heads and the attention is computed for each head independently. To exchange information between heads when the attention is calculated, we propose to overlap $Q$, $K$, $V$ with $Q$, $K$, and $V$ in adjacent heads (right). Since overlapped heads would slightly increase the number of dimensions, the projection matrix would project the concatenated heads to the original token dimension.}
\label{fig:mohsa}
\end{figure*}

\subsection{Multi-Head Self-Attention}

Transformer \cite{vaswani2017attention} has an attention mechanism to compute the long-range relationship between tokens. The attention calculation includes queries, keys, values, and the dimension of queries and keys $d_k$. The matrix format for attention calculation \cite{vaswani2017attention} is illustrated in Eq.~\eqref{eq:attn}.
\begin{equation}
\label{eq:attn}
    Attention(Q, K, V) = Softmax(\frac{QK^T}{\sqrt{d_k}})V
\end{equation}

Multi-Head Self-Attention is utilized in the Transformer model for better performance so that different heads can learn different representations, which is better than one-head attention \cite{vaswani2017attention}. The attention in each head is illustrated in Eq.~\eqref{eq:head}. $Q$, $K$, and $V$ are divided as $Q_i$, $K_i$, and $V_i$ in each head, respectively, and the attention calculation is implemented in each head independently for different aspects of learning.
\begin{equation}
\label{eq:head}
    Head_i = Attention(Q_i, K_i, V_i)
\end{equation}
\begin{equation}
\label{eq:mhsa}
    MHSA(Q, K, V) = Concate(Head_1,...,Head_n)W
\end{equation}

Finally, MHSA could be represented as Eq.~\eqref{eq:mhsa}. The results of all heads are concatenated and $W$ is the projection matrix.

Nevertheless, the attention is implemented independently in each head, and one head does not have the attention information of the other heads when the attention is computed. Although the projection matrix is implemented after all heads are concatenated, only the information in each token is exchanged. There is no information exchange when computing the attention in each head. Thus, we propose to overlap the information of neighboring heads to enhance the information exchange when the attention is computed in each head.

\subsection{Multi-Overlapped-Head Self-Attention}

To facilitate information exchange between heads, we exploit soft division instead of hard division when $Q$, $K$, and $V$ are divided into different heads. The process could be illustrated in Eq.~\eqref{eq:overlaped_qkv} for $Q$, $K$, and $V$ respectively. We utilize ``part" to illustrate partial overlap with adjacent heads in Eq.~\eqref{eq:overlaped_qkv}.
\begin{equation}
\label{eq:overlaped_qkv}
\begin{split}
  &  Q_i' = Concate(part(Q_{i-1}), Q_i, part(Q_{i+1})) \\
  &  K_i' = Concate(part(K_{i-1}), K_i, part(K_{i+1})) \\
  &  V_i' = Concate(part(V_{i-1}), V_i, part(V_{i+1}))
\end{split}
\end{equation}

In Eq.~\eqref{eq:overlaped_qkv}, $Q_i$, $K_i$ and $V_i$ are the original hard division results for $Head_i$, which are shown in Eq.~\eqref{eq:head}. For soft division, $Q_i'$, $K_i'$, and $V_i'$ also include partial information in their neighboring heads so that $Q$, $K$, $V$ are overlapped with two adjacent heads, which is demonstrated in Fig.~\ref{fig:mohsa}. The left figure indicates the original implementation of the hard division of $Q$, $K$, $V$ to different heads, and the right figure represents our proposed implementation of the soft division of $Q'$, $K'$, $V'$ to different heads. The $Q_i$, $K_i$, and $V_i$ would overlap with two adjacent heads to construct $Q_i'$, $K_i'$, and $V_i'$ to calculate the attention. For the first head and the last head, which have only one adjacent head, zero padding is utilized to construct two neighboring heads for the first and the last head.
\begin{equation}
\label{eq:overlaped_head}
    Head_i' = Attention(Q_i', K_i', V_i')
\end{equation}
\begin{equation}
\label{eq:mohsa}
    MOHSA(Q, K, V) = Concate(Head_1',...,Head_n')W'
\end{equation}

After the attention for the overlapped $Q'$, $K'$, and $V'$ is calculated for each head, the results are concatenated together, as illustrated in Eq.~\ref{eq:overlaped_head}-\ref{eq:mohsa}. The overlapped heads would slightly increase the dimension of the tokens after concatenation. Thus, the projection matrix $W'$ would project the concatenated dimension ($h*dim_h'$) to the original token dimension ($dim$) so that the token could be fed into the next layer with the same dimension. The projection matrix $W$ in Eq.~\eqref{eq:mhsa} projects the concatenated non-overlapped heads ($h*dim_h$) to the original token dimension ($dim$). Since the dimension of overlapped heads $dim_h'$ is slightly larger than the dimension of non-overlapped heads $dim_h$ and the number of heads $h$ is unchanged, the projection matrix $W'$ in our proposed MOHSA would have slightly more parameters than the projection matrix $W$ in the original MHSA.

\begin{figure}[t]
\begin{center}
   \includegraphics[width=0.85\linewidth]{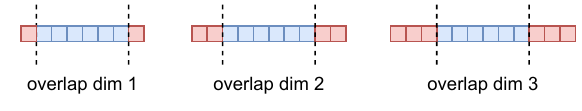}
\end{center}
   \caption{The illustration of the overlap dimensions. The blue parts demonstrate the original non-overlapping heads, and the red parts indicate the overlapped parts from adjacent heads. The number of overlapping dimensions is the overlap dimension of one side of the adjacent head.}
\label{fig:overlap}
\end{figure}

\begin{table}[t]
\centering
\setlength{\tabcolsep}{10pt}
\renewcommand{\arraystretch}{1.0}
\caption{The variants for overlapping ratios}
{\begin{tabular}{c|c}
 \hline
Methods & Description \\
\hline
inc (x layers) & increase overlap dim by 1 every x layers \\
dec (x layers) & decrease overlap dim by 1 every x layers \\
\hline
0-indexed & start (end) overlap dim from (to) 0 \\
1-indexed & start (end) overlap dim from (to) 1 \\

\hline

\end{tabular}
}
\label{table:overlap}
\end{table}

\subsection{Overlapping Ratios}
\label{overlapping_ratios} 
In the experiments, we design several paradigms for the overlapping ratios. In this work, we utilize overlap dimensions to demonstrate overlap ratios, as illustrated in Fig.~\ref{fig:overlap}. The number of overlap dimensions for each head is the overlap dimension of its adjacent head on one side. In Fig.~\ref{fig:overlap}, the blue sections are the original parts of the heads and the red sections are the overlapped parts with two adjacent heads. For the first and the last heads that have only one side adjacent head, zero padding is exploited for the missing neighboring heads.

In addition to the fixed overlap dimensions for all layers, we also implement some variants of the overlapping ratios by changing the overlap dimensions according to the depths of the layers. In this paper, one layer includes the attention module and the FFN module, which represents one layer of the Transformer encoder. The variants for the overlap ratios are demonstrated in Table~\ref{table:overlap}. In Table~\ref{table:overlap}, ``inc (x layers)" represents the overlap dimension is increased by 1 every x layers, and ``dec (x layers)" indicates that the overlap dimension is decreased by 1 every x layers, which is a reverse process of "inc (x layers)" based on the depths of the layers. In addition, ``0-indexed" illustrates that the overlap dimension starts from 0 for ``inc" and ends at 0 for ``dec", and ``1-indexed" demonstrates that the overlap dimension starts from 1 for ``inc" and ends at 1 for ``dec".

\begin{table}[t]
\centering
\setlength{\tabcolsep}{6pt}
\renewcommand{\arraystretch}{1.0}
\caption{The parameter settings for the models}
{\begin{tabular}{c|c|c|c}
 \hline
Models & heads & depths & MLP ratio \\
\hline
ViT-Tiny & 12 & 12 & 4  \\
ViT-Small & 12 & 12 & 4  \\
CaiT-xxs12 & 4 & 12 (2) & 4  \\
CaiT-xxs24 & 4 & 24 (2) & 4 \\
Swin-Tiny & (3, 6, 12, 24) & (2, 2, 6, 2) & 4 \\

\hline

\end{tabular}
}
\label{table:param_models}
\end{table}

\section{Experiments}

In the experiments, we select three representative models to investigate the proposed approach: vanilla ViT \cite{dosovitskiy2020image}, deeper layer Transformer CaiT \cite{touvron2021going}, and window-based hierarchical Transformer Swin-Transformer \cite{liu2021swin}. More specifically, ViT-Tiny \cite{dosovitskiy2020image}, ViT-Small \cite{dosovitskiy2020image}, CaiT-xxs12 \cite{touvron2021going}, CaiT-xxs24 \cite{touvron2021going} and Swin-Tiny \cite{liu2021swin} are selected to explore the effectiveness of the proposed approach. The parameter settings for the models are shown in Table~\ref{table:param_models}. For Swin-Tiny, the number of heads and depths is varied depending on the stages. CaiT-xxs12 and CaiT-xxs24 have two extra class layers. The models will be trained and evaluated on CIFAR-10 \cite{krizhevsky2009learning}, CIFAR-100 \cite{krizhevsky2009learning}, Tiny-ImageNet \cite{le2015tiny} and ImageNet-1k \cite{russakovsky2015imagenet}.

The experiments are implemented by 100 epochs with 20 epochs warmup. The optimizer is AdamW \cite{kingma2014adam}. The experiments for CIFAR-10 \cite{krizhevsky2009learning}, CIFAR-100 \cite{krizhevsky2009learning}, Tiny-ImageNet \cite{le2015tiny} are run on 4 NVIDIA P100 GPUs with total batch size of 128. The experiments for ImageNet-1k \cite{russakovsky2015imagenet} are conducted on 4 NVIDIA V100 GPUs with a total batch size of 512. The initial learning rate for ViT and CaiT is 0.0005. For Swin-Tiny, the initial learning rate for Tiny-ImageNet is 0.00025, and for ImageNet is 0.0005. The images are resized to 224 for the experiments. Some other settings follow the settings in Swin-Transformer \cite{liu2021swin}.

\begin{table}[t]
\centering
\setlength{\tabcolsep}{3pt}
\renewcommand{\arraystretch}{1.0}
\caption{The ablation study on ViT for CIFAR-10}
{\begin{tabular}{c|c|c|c|c|c|c}
 \hline
  \multirow{2}{*}{Methods} & \multicolumn{3}{c|}{ViT-Tiny} & 
     \multicolumn{3}{c}{ViT-Small} \\
 \cline{2-7}
 & Accuracy & Params & FLOPs & Accuracy & Params & FLOPs \\
\hline
original & 84.31 & 5.5M & 1.3G & 86.53 & 21.7M & 4.6G \\
\hline
fixed 1 & 84.94 & 5.6M & 1.3G & 86.98 & 21.8M & 4.7G \\
fixed half & 85.34 & 6.0M & 1.5G & 86.84 & 23.4M & 5.3G \\
\hline
inc-0 (1) & 85.54 & 5.8M & 1.4G & \textbf{87.30} & 22.3M & 4.9G \\
inc-0 (3) & 84.83 & 5.6M & 1.3G & 87.24 & 21.8M & 4.7G \\
inc-0 (6) & \textbf{85.65} & 5.5M & 1.3G & 86.74 & 21.7M & 4.6G \\
\hline
inc-1 (1) & 85.35 & 5.9M & 1.5G & 86.98 & 22.4M & 4.9G \\
inc-1 (3) & 84.31 & 5.7M & 1.3G & 86.83 & 21.9M & 4.7G \\
inc-1 (4) & 85.18 & 5.6M & 1.3G & 86.81 & 21.9M & 4.7G \\
\hline
dec-0 (1) & 85.43 & 5.8M & 1.4G & 86.67 & 22.3M & 4.9G \\
dec-0 (3) & 84.81 & 5.6M & 1.3G & 86.35 & 21.8M & 4.7G \\
\hline
dec-1 (1) & 85.25 & 5.9M & 1.5G & 86.87 & 22.4M & 4.9G \\
dec-1 (3) & 85.12 & 5.7M & 1.3G & 86.84 & 21.9M & 4.7G \\
\hline
\end{tabular}
}
\label{table:vit_cifar10}
\end{table}

\begin{table}[t]
\centering
\setlength{\tabcolsep}{8pt}
\renewcommand{\arraystretch}{1.0}
\caption{The ablation study on CaiT-xxs12 for CIFAR-10}
{\begin{tabular}{c|c|c|c}
 \hline
Methods & Accuracy & Params & FLOPs \\
\hline
original & 77.04 & 6.4M & 1.3G \\
\hline
fixed 1 & 78.14 & 6.4M & 1.3G \\
\hline
inc-0 (1) & 78.79 & 6.5M & 1.4G \\
inc-0 (2) & 78.63 & 6.5M & 1.3G \\
inc-0 (3) & \textbf{80.04 (+3.00)} & 6.4M & 1.3G \\
\hline

\end{tabular}
}
\label{table:cait_cifar10}
\end{table}

\subsection{CIFAR-10}
The ablation study for ViT-Tiny and ViT-Small on CIFAR-10 is demonstrated in Table~\ref{table:vit_cifar10}. In Table~\ref{table:vit_cifar10}, ``fixed 1" represents that the overlap dimension is 1 for all layers, and ``fixed half" indicates that the overlap dimension is half of the head dimension for all layers. More details on variants of overlapping ratios can be found in Sec.~\ref{overlapping_ratios}.

The best accuracy for ViT-Tiny is 85.65 for variant ``inc-0 (6)" and the best accuracy for ViT-Small is 87.30 for variant ``inc-0 (1)". For ViT-Tiny, ``fixed half" demonstrates higher accuracy than ``fixed 1". For ViT-Small, ``fixed half" has lower accuracy than ``fixed 1".

Although the overlapped heads would slightly increase the computations and parameters, from the experimental results, we can see that the increased number of parameters and FLOPs are negligible. Table~\ref{table:cait_cifar10} illustrates the experimental results of CaiT-xxs12 on CIFAR-10. The best result 80.04 significantly boosts the performance of CaiT-xxs12 by 3\% with negligible overhead.

\begin{table}[t]
\centering
\setlength{\tabcolsep}{3pt}
\renewcommand{\arraystretch}{1.0}
\caption{The ablation study on ViT for CIFAR-100}
{\begin{tabular}{c|c|c|c|c|c|c}
 \hline
  \multirow{2}{*}{Methods} & \multicolumn{3}{c|}{ViT-Tiny} & 
     \multicolumn{3}{c}{ViT-Small} \\
 \cline{2-7}
 & Accuracy & Params & FLOPs & Accuracy & Params & FLOPs \\
\hline
original & 61.88 & 5.5M & 1.3G & 63.78 & 21.7M & 4.6G \\
\hline
fixed 1 & 62.17 & 5.6M & 1.3G & 63.90 & 21.8M & 4.7G \\
fixed half & 62.61 & 6.0M & 1.5G & 63.76 & 23.5M & 5.3G \\
\hline
inc-0 (1) & 62.86 & 5.8M & 1.4G & 64.83 & 22.3M & 4.9G \\
inc-0 (2) & \textbf{63.01} & 5.7M & 1.3G & 64.21 & 22.0M & 4.7G \\
\hline
inc-1 (1) & 62.80 & 5.9M & 1.5G & 64.49 & 22.4M & 4.9G \\
inc-1 (2) & 62.77 & 5.7M & 1.4G & 64.24 & 22.1M & 4.8G \\
\hline
dec-0 (1) & 62.45 & 5.8M & 1.4G & 64.62 & 22.3M & 4.9G \\
\hline
dec-1 (3) & 62.25 & 5.7M & 1.3G & \textbf{64.97} & 22.0M & 4.7G \\
\hline
\end{tabular}
}
\label{table:vit_cifar100}
\end{table}

\subsection{CIFAR-100} 
Table~\ref{table:vit_cifar100} demonstrates the experiments of ViT-Tiny and ViT-Small on CIFAR-100. The best result for ViT-Tiny is 63.01 with ``inc-0 (2)" and for ViT-small is 64.97 with ``dec-1 (3)". Both results improve the models by more than 1\% over the original models. Similar to the results on CIFAR-10, ``fixed half" has higher accuracy for ViT-Tiny, and ``fixed 1" illustrates better performance for ViT-Small. In addition, ``fixed half" does not demonstrate effectiveness for ViT-Small.

Table~\ref{table:cait_cifar100} illustrates the performance of CaiT on CIFAR-100. For CaiT-xxs12, using 1 overlap dimension for all layers could boost the accuracy by nearly 2\%. For CaiT-xxs24, the accuracy is significantly enhanced by 5\% with ``inc-1 (3)". Additionally, ``inc-1 (3)" is much better than using the half dimension of the head as the overlap dimension, which utilizes more parameters and FLOPs.

\begin{table}[t]
\centering
\setlength{\tabcolsep}{6pt}
\renewcommand{\arraystretch}{1.0}
\caption{The ablation study on CaiT for CIFAR-100}
{\begin{tabular}{c|c|c|c|c}
 \hline
Models & Methods & Accuracy & Params & FLOPs \\
\hline
\multirow{2}{*}{CaiT-xxs12} & original & 50.95 & 6.4M & 1.3G \\
\cline{2-5}
& fixed 1 & \textbf{52.94 (+1.99)} & 6.4M & 1.3G \\
\cline{2-5}
\hline

\multirow{4}{*}{CaiT-xxs24} & original & 50.94 & 11.8M & 2.5G \\
\cline{2-5}
& fixed half & 52.47 & 12.7M & 3.1G \\
\cline{2-5}
& inc-1 (1) & 55.84 & 12.2M & 2.8G \\
& inc-1 (3) & \textbf{55.94 (+5.00)} & 11.9M & 2.6G \\

\hline
\end{tabular}
}
\label{table:cait_cifar100}
\end{table}

\subsection{Tiny-ImageNet}

\begin{table}[h!]
\centering
\caption{The experimental results on Tiny-ImageNet}
\renewcommand{\arraystretch}{1.0}
\setlength{\tabcolsep}{6pt}
\begin{tabular}{c|c|c|c|c}
 \hline
Models & Methods & Accuracy & Params & FLOPs  \\
\hline
\multirow{7}{*}{ViT-Tiny} & original & 50.79 & 5.6M & 1.3G \\ 
\cline{2-5}

& fixed 1 & 51.44 & 5.6M & 1.3G \\
& fixed half & 52.32 & 6.0M & 1.5G \\ 
\cline{2-5}

& inc-0 (1) & 51.45 & 5.9M & 1.4G \\
& inc-1 (2) & \textbf{52.40 (+1.61)} & 5.7M & 1.4G \\ 
\cline{2-5}

& dec-0 (1) & 52.21 & 5.9M & 1.4G \\
& dec-1 (2) & 52.20 & 5.7M & 1.4G \\
\hline

\multirow{6}{*}{ViT-Small} & original & 54.53 & 21.7M & 4.6G \\
\cline{2-5}

& fixed 1 & 55.15 & 21.8M & 4.7G \\
& fixed half & 54.65 & 23.5M & 5.3G \\
\cline{2-5}

& inc-0 (2) & 55.30 & 22.0M & 4.7G \\
& inc-1 (1) & \textbf{55.80 (+1.27)} & 22.4M & 4.9G \\
\cline{2-5}

& dec-1 (3) & 55.46 & 22.0M & 4.7G \\

\hline

\multirow{2}{*}{CaiT-xxs12} & original & 42.66 & 6.5M & 1.3G \\
\cline{2-5}

& fixed 1 & \textbf{45.05 (+2.39)} & 6.5M & 1.3G \\
\hline

\multirow{3}{*}{CaiT-xxs24} & original & 42.46 & 11.8M & 2.5G \\
\cline{2-5}

& fixed 1 & 46.04 & 11.8M & 2.6G \\
\cline{2-5}

& inc-0 (2) & 46.98 & 12.0M & 2.7G \\
& inc-1 (1) & \textbf{49.87 (+7.41)} & 12.2M & 2.8G \\
\hline

\multirow{6}{*}{Swin-Tiny} & original & 57.04 & 27.7M & 4.5G \\
\cline{2-5}
& fixed 1 & 58.27 & 27.8M & 4.5G \\
\cline{2-5}

& inc-0 (6) & 58.67 & 27.8M & 4.5G \\
& inc-1 (3) & 58.58 & 28.1M & 4.6G \\
\cline{2-5}

& dec-0 (1) & 58.59 & 28.0M & 4.7G \\
& dec-1 (1) & \textbf{58.90 (+1.86)} & 28.1M & 4.7G \\
\hline

\end{tabular}
\label{table:tiny_imagenet}
\end{table}

The experimental results on Tiny-ImageNet are demonstrated in Table~\ref{table:tiny_imagenet}. The accuracy of ViT-Tiny could be enhanced by 1.61\% utilizing ``inc-1 (2)". ViT-Small is boosted by 1.27\% with ``inc-1 (1)". Moreover, using half of the head dimension as the overlap dimension is better than using 1 as the overlap dimension for all layers for ViT-Tiny, while it is not the case for ViT-Small.

For CaiT-xxs12 and CaiT-xxs24, ``fixed 1" that utilizes a fixed overlap dimension of 1 for all layers could significantly improve the accuracy by 2.39\% and 3.58\%, respectively. Furthermore, the accuracy is enormously enhanced by 7.41\% for CaiT-xxs24 with ``inc-1 (1)". The effectiveness of our method is significant on CaiT models and we can ignore the slightly increased parameters and computations.

For Swin-Tiny, ``fixed 1" paradigm could greatly enhance the accuracy by 1.23\%, and the performance could be further boosted by a large margin with ``dec-1 (1)". The effectiveness of overlapped heads is also demonstrated on the Transformer model with window-based hierarchical architecture.

\subsection{ImageNet}
The experimental results on ImageNet are demonstrated in Table~\ref{table:imagenet}. Almost all models equipped with our proposed MOHSA have significant improvements on ImageNet.

\begin{table}[h!]
\centering
\caption{The experimental results on ImageNet}
\setlength{\tabcolsep}{6pt}
\renewcommand{\arraystretch}{0.95}
\begin{tabular}{c|c|c|c|c}
 \hline
Models & Methods & Accuracy & Params & FLOPs  \\
\hline
\multirow{7}{*}{ViT-Tiny} & original & 69.25 & 5.7M & 1.3G \\ 
\cline{2-5}

& fixed 1 & 69.11 & 5.8M & 1.3G \\
& fixed half & \textbf{70.31 (+1.06)} & 6.2M & 1.5G \\
\cline{2-5}

& inc-0 (1) & 69.96 & 6.0M & 1.4G \\
& inc-1 (1) & 69.70 & 6.1M & 1.5G \\ 
\cline{2-5}

& dec-0 (1) & 69.71 & 6.0M & 1.4G \\
& dec-1 (1) & 69.34 & 6.1M & 1.5G \\
\hline

\multirow{7}{*}{ViT-Small} & original & 77.24 & 22.0M & 4.6G \\
\cline{2-5}

& fixed 1 & 77.57 & 22.1M & 4.7G \\
& fixed half & 77.40 & 23.8M & 5.3G \\
\cline{2-5}

& inc-0 (1) & 77.87 & 22.6M & 4.9G \\
& inc-1 (1) & 77.91 & 22.8M & 4.9G \\ 
\cline{2-5}

& dec-0 (1) & \textbf{78.03 (+0.79)} & 22.6M & 4.9G \\
& dec-1 (1) & 77.99 & 22.8M & 4.9G \\
\hline

\hline

\multirow{7}{*}{CaiT-xxs12} & original & 67.56 & 6.6M & 1.3G \\
\cline{2-5}

& fixed 1 & 67.61 & 6.6M & 1.3G \\
& fixed half & 68.30 & 7.0M & 1.6G \\
\cline{2-5}

& inc-0 (1) & 68.63 & 6.7M & 1.4G \\
& inc-1 (1) & \textbf{68.73 (+1.17)} & 6.7M & 1.4G \\ 
\cline{2-5}

& dec-0 (1) & 68.49 & 6.7M & 1.4G \\
& dec-1 (1) & 68.25 & 6.7M & 1.4G \\
\hline

\multirow{7}{*}{CaiT-xxs24} & original & 70.56 & 11.9M & 2.5G \\
\cline{2-5}

& fixed 1 & 73.64 & 12.0M & 2.6G \\
& fixed half & 73.79 & 12.8M & 3.1G \\
\cline{2-5}

& inc-0 (1) & \textbf{74.26 (+3.70)} & 12.4M & 2.8G \\
& inc-1 (1) & 74.19 & 12.4M & 2.8G \\ 
\cline{2-5}

& dec-0 (1) & 74.08 & 12.4M & 2.8G \\
& dec-1 (1) & 74.17 & 12.4M & 2.8G \\
\hline

\multirow{7}{*}{Swin-Tiny} & original & 78.52 & 28.3M & 4.5G \\
\cline{2-5}
& fixed 1 & 78.44 & 28.4M & 4.5G \\
& fixed half & 78.70 & 30.4M & 5.0G \\
\cline{2-5}

& inc-0 (1) & 78.70 & 29.4M & 4.7G \\
& inc-1 (1) & \textbf{78.72 (+0.20)} & 29.6M & 4.7G \\ 
\cline{2-5}

& dec-0 (1) & 78.64 & 28.6M & 4.7G \\
& dec-1 (1) & 78.61 & 28.8M & 4.7G \\
\hline

\end{tabular}
\label{table:imagenet}
\end{table}

ViT-Tiny could be boosted by more than 1\% with ``fixed half" that uses half of the head dimension as the overlap dimension for all layers. ViT-Small is enhanced by 0.79\% with ``dec-0 (1)" with negligible overhead. Additionally, utilizing 1 as the overlap dimension for all layers is not effective for ViT-Tiny on ImageNet. But using 1 as the overlap dimension is better than using half head dimension as the overlap dimension for all layers for ViT-Small.

Our approach has even more performance enhancement on CaiT models. CaiT-xxs12 is greatly improved by 1.17\% with ``inc-1 (1)" and the accuracy of CaiT-xxs24 is significantly increased by 3.70\% with ``inc-0 (1)". For CaiT-xxs24, the accuracy is boosted by more than 3\% by using only 1 as the overlap dimension for all layers, which illustrates the effectiveness and efficiency of our proposed method by remarkably boosting the performance of the models with minimal overhead.

\begin{table}[t]
\centering
\setlength{\tabcolsep}{3pt}
\renewcommand{\arraystretch}{1.0}
\caption{The ablation study on applying overlap to $Q$, $K$, $V$}
{\begin{tabular}{c|c|c|c|c|c|c}
 \hline
  \multirow{2}{*}{Overlap} & \multicolumn{3}{c|}{ViT-Tiny} & 
     \multicolumn{3}{c}{CaiT-xxs12} \\
 \cline{2-7}
 & Accuracy & Params & FLOPs & Accuracy & Params & FLOPs \\
\hline
None & 69.25 & 5.7M & 1.3G & 67.56 & 6.6M & 1.3G \\
\hline
$Q$, $K$ & 69.15 & 5.7M & 1.3G & 66.85 & 6.6M & 1.3G \\
$V$ & 69.62 & 6.0M & 1.4G & 68.53 & 6.7M & 1.3G \\
$Q$, $K$, $V$ & \textbf{69.96} & 6.0M & 1.4G & \textbf{68.63} & 6.7M & 1.4G \\
\hline
\end{tabular}
}
\label{table:ablation_qkv}
\end{table}

For Swin-Tiny, simply using 1 as the overlap dimension is not effective on ImageNet. Even though increasing the overlap dimension to half of the head dimension improves the accuracy of Swin-Tiny, it achieves almost the same result as the paradigm of varying the overlap dimension by the depths of the layers which have fewer parameters and computations.

Table~\ref{table:ablation_qkv} illustrates the ablation study by applying the overlap paradigm ``inc-0 (1)" to $Q$, $K$, or $V$ using ViT-Tiny and CaiT-xxs12 on ImageNet. From the experimental results shown in Table~\ref{table:ablation_qkv}, applying our approach to $V$ has more effect on the performance improvement, and applying the overlapping approach to $Q$, $K$, and $V$ could yield the best performance.

\begin{figure}[t]
\centering
\includegraphics[width=.15\textwidth]{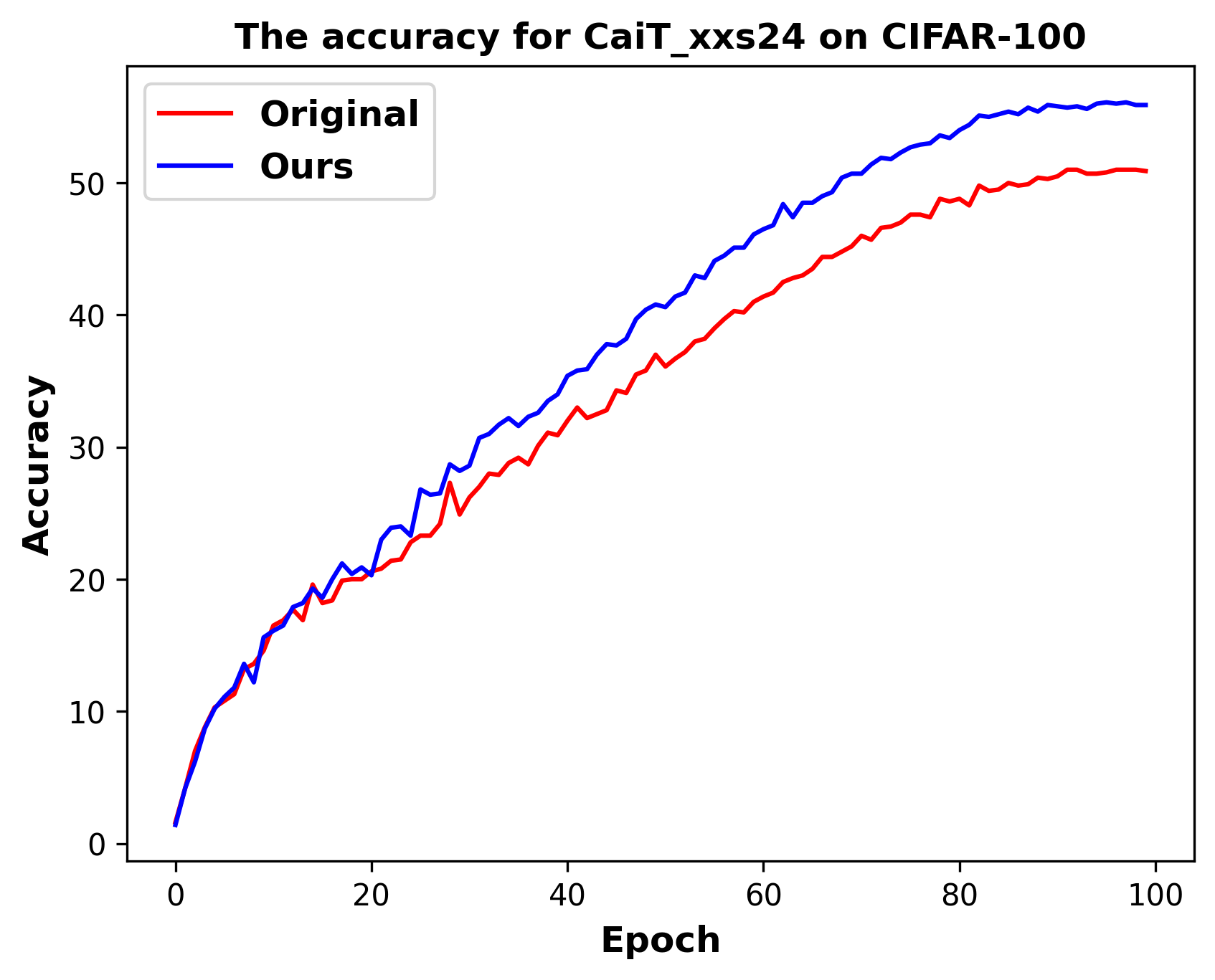}
\includegraphics[width=.15\textwidth]{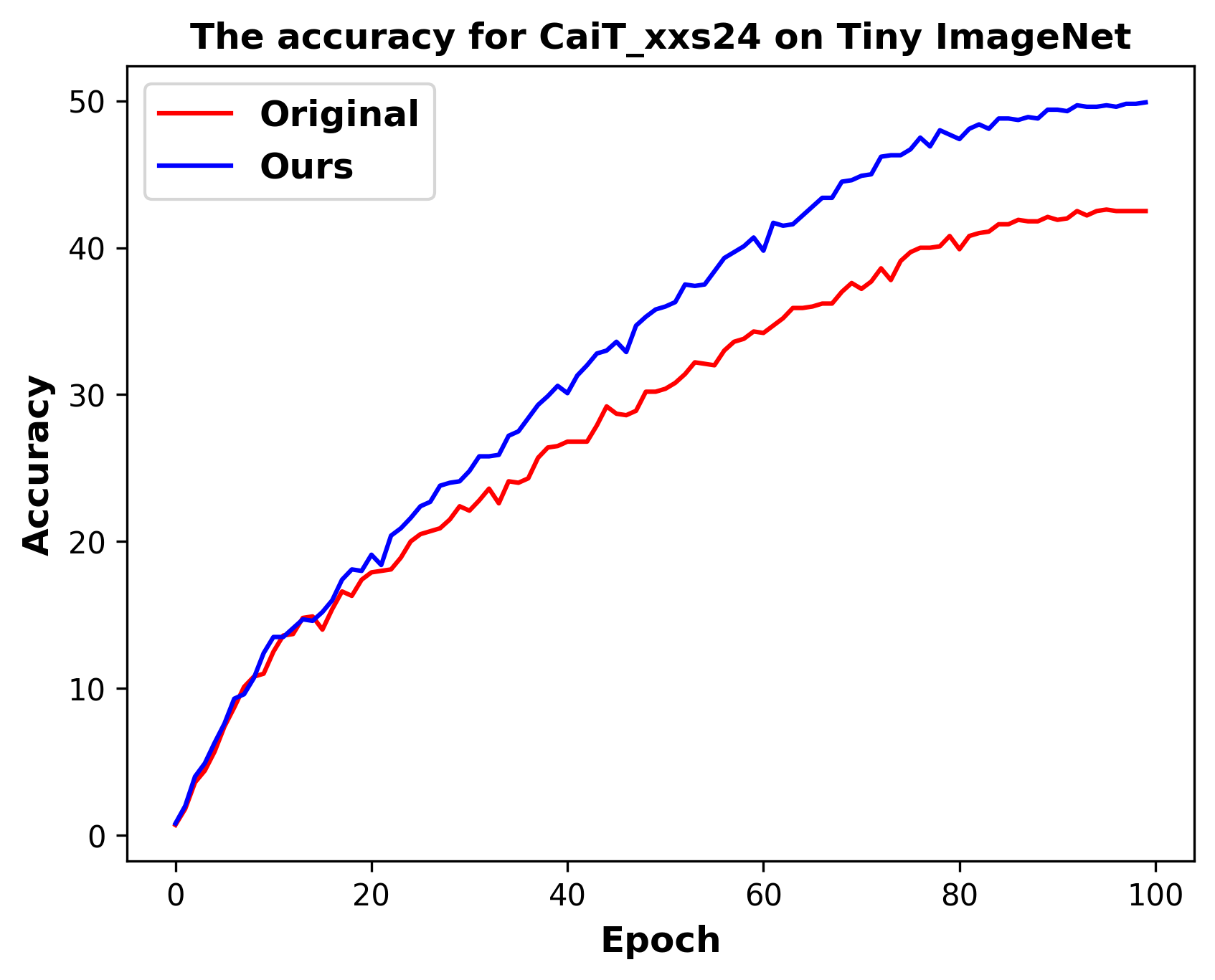}
\includegraphics[width=.15\textwidth]{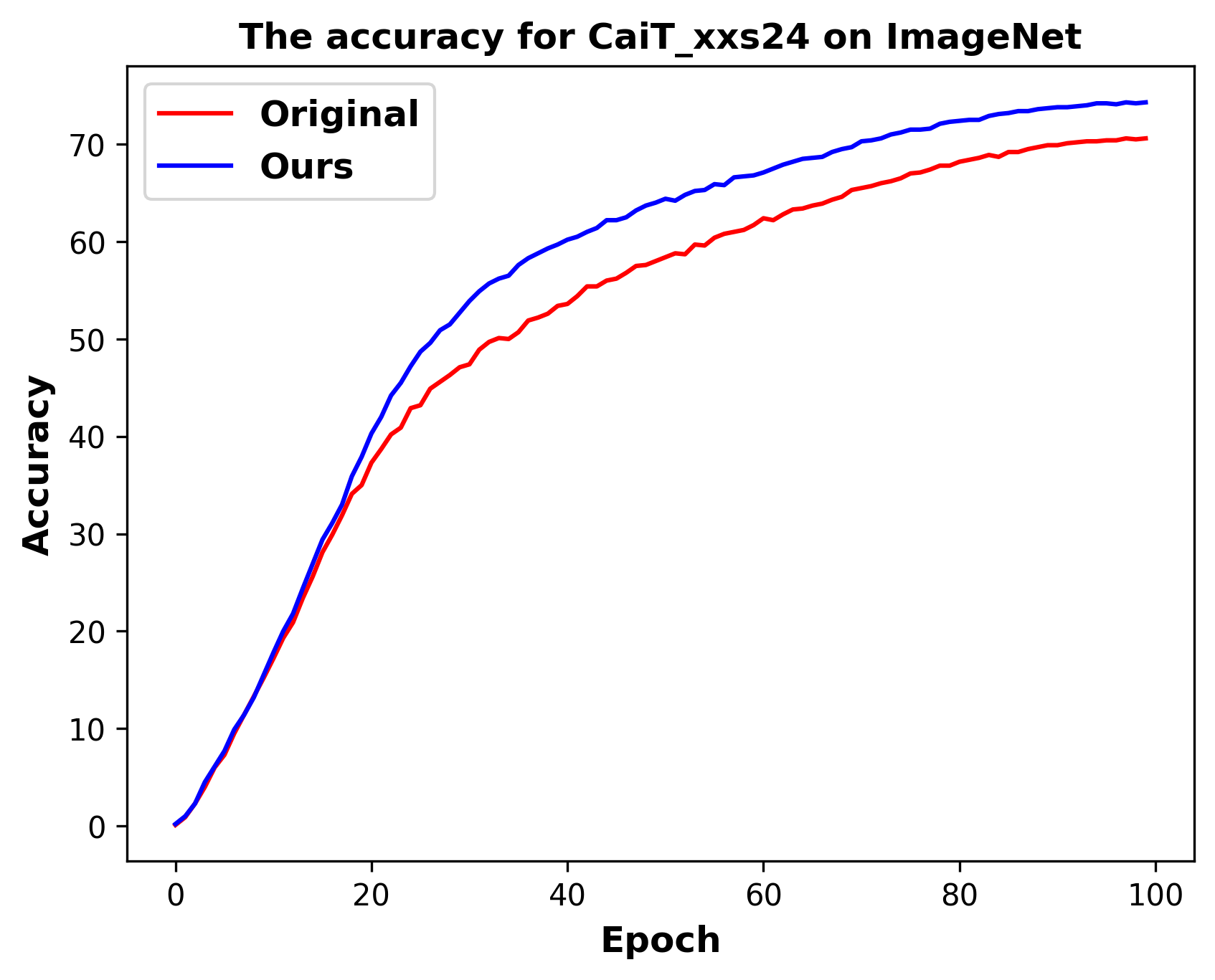}\\
\caption{The accuracy comparison during the training process. The accuracy comparison between the original method and our proposed approach for CaiT-xxs24 on val or test set of CIFAR-100, Tiny-ImageNet, and ImageNet is illustrated from left to right. The blue curve demonstrates our approach with the best performance, and the red curve indicates the original method.}
\label{fig:accuracy_plot}
\end{figure}

\subsection{Analysis}
For different variants of our proposed method, the results manifest some differences. For a fixed paradigm where the overlap dimension is the same for all layers, the performance could be enhanced by only 1 overlap dimension in most cases, and increasing the overlap dimension for fixed mode cannot guarantee better results. Additionally, variants with varying overlap dimensions based on the depths of the layers demonstrate performance superior to the fixed paradigm on various models and datasets in most cases. Compared to the fixed paradigm with a high overlap dimension, varying overlap dimensions with the depths of the layers could save the number of parameters and computational costs. 

To illustrate the accuracy during training, Fig.~\ref{fig:accuracy_plot} manifests the accuracy of CaiT-xxs24 on the val or test set with the training proceeding between the original models and our proposed models. ``inc-1 (3)", ``inc-1 (1)" and ``inc-0 (1)" are selected as ours for comparison for CIFAR-100, Tiny-ImageNet, and ImageNet, respectively. The accuracy between the original models and our models demonstrates almost no difference at the early stage of training, while diverging significantly as the training proceeds.

\section{Conclusion}

In this paper, we have proposed a simple yet effective module MOHSA, to improve the original MHSA in Vision Transformers by overlapping the heads, allowing for information exchange during attention calculation in each head. Extensive evaluations and comparisons with the state-of-the-art on multiple datasets have demonstrated the superior performance of the proposed approach. To the best of our knowledge, this is the first work to propose overlapping heads and achieve significant enhancement across various datasets for different Vision Transformer models. We hope that our work will inspire the community to further explore the structure of Vision Transformers.

\section*{Acknowledgment}
This research was funded by the Natural Sciences and Engineering Research Council of Canada (NSERC) under grant number RGPIN-2026-05645. Bo Luo was supported in part by US NSF IIS-2014552, DGE-1565570, and the Ripple University Blockchain Research Initiative.

\bibliographystyle{IEEEtran}
\bibliography{reference}

\end{document}